\documentclass[10pt,twocolumn,letterpaper]{article}

\usepackage{wacv} % Camera-ready version
\usepackage{amsmath,amssymb,graphicx,booktabs,url}
\usepackage{stfloats}

\makeatletter
\g@addto@macro\@maketitle{% First-page visual summary, inserted below the unchanged WACV title block.
\begin{center}
\includegraphics[width=0.75\textwidth]{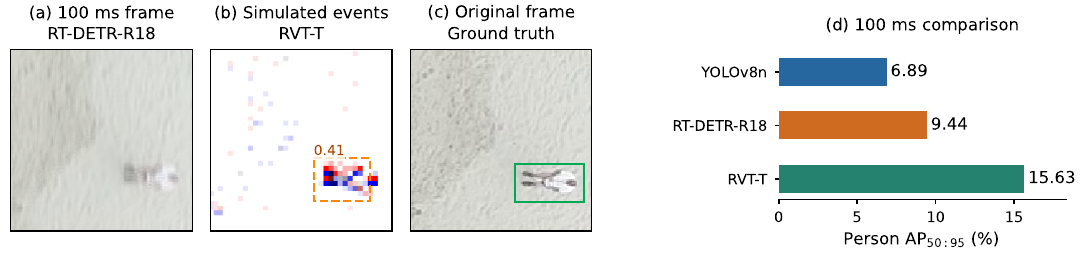}
\captionsetup{hypcap=false}
\captionof{figure}{Person detection under exposure blur in maritime UAV imagery.
The paired example shows (a) a miss by RT-DETR-R18 on the 100\,ms exposure image, (b) a detection by RVT-T using simulated events, and (c) the original target with its ground-truth box.
Green denotes ground truth; dashed orange denotes a prediction at confidence $\geq0.25$.
Red and blue indicate positive and negative events, dark pixels indicate overlapping polarities, and white indicates no events. 
(d) Person AP$_{50:95}$ (\%) on all 843 test samples compares 100\,ms exposure-matched frame detectors with RVT-T using events from the original video.}
\label{fig:teaser}
\end{center}
}
\makeatother

\definecolor{wacvblue}{rgb}{0.21,0.49,0.74}
\usepackage[breaklinks,colorlinks,allcolors=wacvblue]{hyperref}

\def\wacvPaperID{7} % *** Enter the WACV Paper ID here
\def\confName{WACV}
\def\confYear{2027}

\title{Event-Based Person Detection for Maritime UAV Search}

\author{Zijie Ning \qquad Patrick Vandewalle\\
EAVISE-PSI, ESAT, KU Leuven\\
Leuven, Belgium\\
{\tt\small \{zijie.ning,patrick.vandewalle\}@kuleuven.be}
}

\begin{document}
\maketitle
% !TeX root = ../main.tex
\begin{abstract}
Maritime search and rescue requires locating people in the water quickly, motivating the use of UAVs to survey the search area.
During flight, camera motion and the longer exposures used in low light can blur small targets and hinder person detection.
Event cameras record brightness changes as they occur, reducing the motion blur associated with longer exposure times.
We investigate their use for maritime person detection by generating synthetic events using the v2e event simulator from public MOBDrone rescue videos and retaining the existing annotations.
We compare the recurrent vision transformer model RVT-T with YOLOv8n and RT-DETR-R18 trained on exposure-matched images.
On 843 test samples from held-out recordings, RVT-T achieves 15.63\% person AP$_{50:95}$, compared with 6.89\% for YOLOv8n and 9.44\% for RT-DETR-R18 under 100\,ms synthetic exposure.
Increasing exposure times from 50 to 200\,ms progressively reduces frame-based detection accuracy, widening the gap to the event model.
To evaluate the feasibility of on-board processing, we deploy our detector on a Jetson Orin Nano platform.
The RVT-T model retains its accuracy and requires approximately 136\,ms and 2.02\,J per sample, allowing real-time processing on the UAV.

\end{abstract}

% !TeX root = ../main.tex
\section{Introduction}
% 海上救援的任务定义和日益增长的需求
% 海上活动日益增多，面对人类落水的风险，海岸救援组织越来越多使用无人机辅助搜索落水的人类。这是一份与时间赛跑的任务，无人机飞的越快，就越有可能更快发现落水人员。而搜救活动可能跨越昼夜，在光线不足的情况下，延长的曝光时间会带来动态模糊，影响了目标检测模型的表现。事件相机是一种近些年才得到关注的新型传感器，它能够对光线的变化产生反应，不会受到动态模糊的影响。其高动态范围的优势使其也可以在有强烈反光的水面场景发挥作用。
% 说明为什么使用UAV，定义海上救援的困难
%↓
%Why Event
%↓
%为什么要嵌入式部署
%↓
%我们的主要贡献：
% 基于已有的MOBDrone建立了一个synthetic事件数据集 E-MOBDrone
% 验证了事件相机对于海上救援场景的可行性
% 证明了嵌入式部署的可行性
Maritime search and rescue brings together the urgency of locating a person in distress and the difficulty of searching a large expanse of water.
Camera-equipped UAVs can support this task by providing an aerial view and helping rescue teams locate people who have fallen overboard~\cite{mobdrone}.
Their usefulness depends on both how quickly they can survey the search area and how reliably a person can be recognized in the captured observations.
Faster flight could shorten the time needed to cover a search area, making reliable detection during motion an important capability for aerial rescue systems.

Maintaining reliable detection is difficult when people occupy only a few pixels against the water background.
Low illumination and rapid camera motion further challenge conventional imaging.
In low light, increasing exposure time collects more light, but also allows motion to spread a target's appearance over more of the image, resulting in a blurred image~\cite{nighttimeevents}.
At a given exposure time, faster image motion likewise increases blur.
These effects connect the practical goals of searching in low light and flying faster to a common visual problem: preserving enough target detail for person detection.

Event cameras offer an alternative way to capture this detail.
Instead of accumulating a complete image over an exposure, they report brightness changes at individual pixels together with their timestamps~\cite{eventsurvey}.
Their high temporal resolution reduces motion blur, while their wide dynamic range supports imaging across large brightness variations~\cite{eventsurvey}.
Research on nighttime event imaging also demonstrates their potential for capturing motion in low-light scenes~\cite{nighttimeevents}.
For maritime search, these properties motivate an event-based view of the visibility challenges faced by aerial cameras, with the prospect of supporting faster search and operation under difficult illumination.

For onboard use, detection performance must be considered alongside computing and power requirements.
Prior maritime vision studies have examined how detector compression and hardware adaptation trade accuracy for inference efficiency~\cite{rizk2022marineedge,heller2022marineedge}.
Bringing event-based vision into this setting calls for a corresponding understanding of its detection performance and embedded processing cost.
Evaluating both connects the promise of a different sensing approach to the hardware available on a search UAV.

In this work, we explore event-based vision as a sensing approach for person detection in maritime UAV search.
Public MOBDrone rescue footage and its existing annotations provide a common basis for evaluating event-based and frame-based person detection~\cite{mobdrone}.
Using simulated events derived from this footage, we study detection under exposure blur (Fig.~\ref{fig:teaser}) and assess the onboard processing requirements through measurements on an Orin Nano.
The study connects an application-level motivation for event cameras with detection evidence and practical deployment considerations.

Our contributions are as follows:
\begin{itemize}
\item We establish a reproducible MOBDrone-derived evaluation protocol that pairs simulated events and exposure-integrated images while preserving source separation and annotation times.
\item We characterize the person-detection benefit of an event-based system under synthetic exposure blur, including its sensitivity to exposure duration and its associated false positives.
\item We measure the trained detectors' accuracy, latency, and energy consumption on a Jetson Orin Nano to characterize the tradeoff between person-detection performance and onboard processing cost.
\end{itemize}

% !TeX root = ../main.tex
\section{Related Work}
\subsection{Event-based vision}
Event-based vision offers an alternative way to observe moving objects under challenging imaging conditions.
An event camera records brightness changes independently at each pixel, reporting their timing rather than collecting a complete image over a shared exposure~\cite{eventsurvey}.
Its high temporal resolution helps preserve motion detail, while its wide dynamic range accommodates large differences in scene brightness~\cite{eventsurvey}.
These properties are relevant to maritime monitoring from moving UAVs, where small targets must be distinguished against water backgrounds with changing illumination and reflections.

Simulators make event data available for exploring such applications using rendered scenes or existing videos, partially compensating for the limited availability of annotated event data.
ESIM generates events through adaptive scene rendering~\cite{esim}, while video-to-event methods allow annotated video datasets to be reused for event-based learning~\cite{videotoevents}.
The v2e toolbox converts video frames into events with models of sensor response and noise~\cite{v2e}.
This provides a practical route for maritime researchers to build on existing footage and object annotations.
Studies of synthetic-to-real learning further demonstrate transfer to real events for object recognition and semantic segmentation~\cite{videotoevents}, with DA4Event improving transfer for recognition through domain adaptation~\cite{da4event}.

Event-based object detection has already produced successful examples in automotive perception.
The Gen1 and 1 Megapixel (1Mpx) benchmarks provide large-scale recordings with annotations for road users, supporting the development and comparison of learned detectors~\cite{gen1,red}.
The Recurrent Event-camera Detector (RED) uses memory to combine information over time and achieves detection performance comparable to frame-based models on 1Mpx~\cite{red}.
Recurrent Vision Transformers (RVT) combine spatial attention with temporal memory to improve the balance between detection accuracy and processing time~\cite{rvt}.
Later work addresses practical requirements: SAST reduces computation by selecting informative regions, while LEOD makes better use of limited annotations through self-training~\cite{sast,leod}.
Together, these data-generation tools and established detectors provide a starting point for bringing event-based vision to maritime monitoring.
We build on v2e and the Tiny variant of RVT (RVT-T)~\cite{rvt} to investigate person detection in aerial rescue footage and its embedded processing requirements.

\subsection{Maritime object detection}
Public aerial datasets support the development of visual detection for maritime search and rescue.
MOBDrone provides drone footage with annotations for people in the water and other maritime objects~\cite{mobdrone}.
On SeaDronesSee~\cite{seadronessee}, YOLOv7-Sea adapts a frame-based detector to maritime UAV imagery through an additional small-object prediction head and an attention module~\cite{yolov7sea}.
LCSC-UAVNet likewise studies lightweight small-object detection from a maritime UAV perspective~\cite{lcscuavnet}.
These studies connect detector design to the target scales and image content encountered in aerial maritime observation.

Event-based perception has also been explored for marine objects.
Aelmore et al.~\cite{aelmore} combine event filtering, clustering, and motion analysis for vessel detection and tracking.
Marine Event Vision uses event-based generalized Hough-transform template matching and clustering to detect objects on the water surface and underwater~\cite{marineeventvision}.
These applications motivate further exploration of event sensing for maritime vision.
Our study focuses on aerial person detection and examines how exposure blur affects the comparison between event-based and frame-based systems.

\subsection{Embedded maritime vision}
Embedded evaluations connect maritime perception algorithms to the computing resources available on their platforms.
Heller et al.~\cite{heller2022marineedge} study YOLO-based marine object detection and its optimization for several edge devices.
Rizk et al.~\cite{rizk2022marineedge} examine pruning and quantization for human detection in marine environments.
For fluvial navigation, Lambert et al.~\cite{lambert2023deployability} compare segmentation networks and evaluate their performance on a Jetson Nano, relating scene understanding to inference speed on embedded hardware.
Together, these studies motivate evaluating visual accuracy alongside actual device performance.
We apply this perspective to event-based aerial person detection by measuring the trained models' accuracy, latency, and energy on an Orin Nano.

% !TeX root = ../main.tex
\section{Methodology}

\begin{figure*}[t]
\centering
\includegraphics[width=\textwidth]{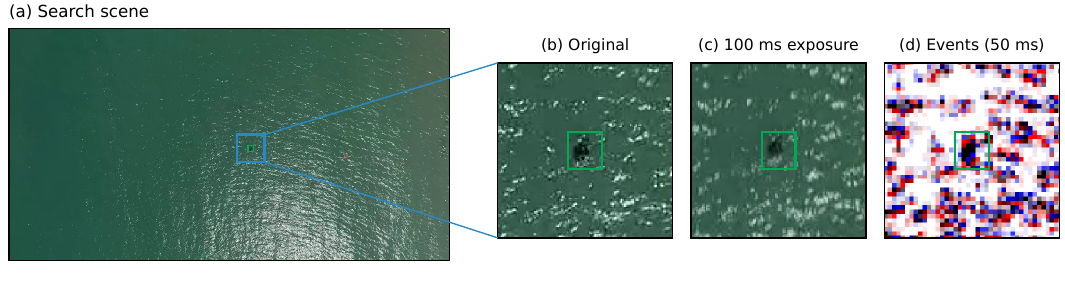}
\caption{A small person in a maritime UAV search scene and three paired views of the same region.
The blue rectangle in (a) locates the region shown at equal scale in (b) the original frame, (c) the 100\,ms exposure image, and (d) the simulated event view.
Green boxes mark the selected person's ground truth.
Events visualize the final 50\,ms (red: increasing brightness; blue: decreasing brightness; dark: overlapping polarities; white: no events), while RVT-T processes a 200\,ms history.}
\label{fig:scene-context}
\end{figure*}

\subsection{Event Data}
An event stream is a sequence of brightness-change measurements, each represented as $e=(x,y,\tau,p)$~\cite{eventsurvey}.
The coordinates $(x,y)$ locate the responding pixel, $\tau$ records the time of the change, and the polarity $p$ indicates whether brightness increased or decreased.
Each pixel triggers an event when its log intensity changes by a contrast threshold relative to its stored reference~\cite{eventsurvey}.
The reference is updated after each event, allowing the pixel to report subsequent changes independently.
Positive (ON) and negative (OFF) events describe increases and decreases in brightness, respectively.
As a person's boundary moves across the image, successive pixels respond to the contrast between the person and the surrounding water.
UAV motion can produce this response even for a stationary target, while waves and changing reflections generate additional events in the background.
Together, these measurements describe where and when changes occur as the camera and objects move.

Our red-and-blue event views accumulate these measurements over a short time window: red indicates positive events, blue indicates negative events, and white indicates no events; dark pixels indicate overlapping positive and negative activity.
To use this information for detection, we collect events over a recent time interval and organize them into spatial maps of event counts, separated by time and polarity.

\subsection{Dataset}
We use MOBDrone because it provides continuous aerial rescue footage and existing object annotations, supporting both video-to-event conversion and exposure formation from the same scenes~\cite{mobdrone}.
The dataset contains 66 video fragments with a native resolution of $1920\times1012$.
We retain all five annotated categories: person, boat, surfboard, wood, and lifebuoy, with person detection as the primary application task.
The person category includes people in the water and on boats or boards.
To evaluate generalization across recordings, we assign original video identities to disjoint training, validation, and test sets, keeping all fragments and observation variants from each source together.
We define a sample as one annotated time together with its corresponding frame observation or event history and object labels.
Sampling every 15th source frame (approximately every 0.5 s at approximately 29.97 frames/s) gives 5,839 training, 1,691 validation, and 843 test samples.
The test set contains five original sources videos and 835 person instances.
We retain its 392 samples without annotated persons to assess false positives alongside person detection.

The small size of people in the aerial view motivates evaluating localization at the original image resolution.
Of the test person boxes, 419 fall within the COCO small-object area range~\cite{coco}.
Figure~\ref{fig:scene-context} illustrates the target scale in a complete search image: the selected person occupies a $23.3\times25.8$-pixel annotation box against the surrounding water.

% Scene figure declared at the start of this section to keep it near the dataset description.

\subsection{Data Preparation}
\paragraph{Paired observations.}
To compare sensing alternatives on the same search scenes, we construct frame observations and event histories at the same annotation times.
Decoded video timestamps align the observations with the original labels.
We scale both spatial coordinates by $1/3$ and add black padding at the bottom to obtain a $640\times640$ input canvas.
This transformation gives the models a common input geometry while preserving the correspondence with native-coordinate boxes.
Frames use bilinear affine resampling, which at this exact scale samples every third native pixel without antialiasing.
Event counts are summed over $3\times3$ native-pixel cells.
At this scale, the example person in Fig.~\ref{fig:scene-context} occupies approximately $7.8\times8.6$ pixels.

\paragraph{Frame observation conditions.}
We use original frames as the reference condition and construct reduced-brightness and exposure-integrated observations to examine the effects of brightness and motion accumulation.
The reduced-brightness condition multiplies encoded pixel values by 0.25, rounds to the nearest integer, and stores the result as an 8-bit image.
To study motion during exposure, we integrate linearized RGB radiance $L(s)$ over an interval of duration $T$ ending at annotation time $t$:
\begin{equation}
 B_t=\Gamma\left(\frac{1}{T}\int_{t-T}^{t} L(s)\,ds\right),
\end{equation}
where $\Gamma$ converts the integrated radiance back to encoded RGB.
We approximate the integral by piecewise-linear interpolation between decoded video timestamps, using observations up to $t$.
The label describes the person's position at the end of the interval, giving the detector the task of localizing the current position from the accumulated appearance.
We chose 100\,ms as the primary exposure after inspecting training examples for visible person blur and interpretable boxes.
Additional 50 and 200\,ms conditions examine how the comparison changes with integration duration.

\paragraph{Event generation.}
To provide an event-based observation of the same scenes, we apply v2e to the original video at its source cadence~\cite{v2e}.
The simulator generates an event when the change in log intensity at a pixel reaches its contrast threshold.
We use a threshold of 0.2 with threshold variability of 0.03, an unlimited photoreceptor bandwidth, and disabled shot, leak, and photoreceptor noise.
For the reduced-brightness condition, we regenerate events from the brightness-scaled video.
For the exposure comparisons, events from the original video provide a shared reference, and the corresponding RVT-T entries reuse the same predictions across integration durations.

\paragraph{Event representation.}
For each sample, we collect the 200\,ms of events leading up to its annotation time and divide this history into four consecutive 50\,ms windows.
Within each window, we count events at each spatial location in ten equal 5\,ms time bins, keeping positive and negative polarities separate.
Each bin therefore contributes two spatial maps: one for increases in brightness and one for decreases.
Stacking these maps gives twenty channels on the common $640\times640$ canvas for each window.
The resulting sample contains a sequence of four twenty-channel representations, preserving both the locations of brightness changes and their order over time.
For visualization, we sum the time bins of the final 50\,ms window and display positive counts in red and negative counts in blue.
The detector processes the full sequence of four representations.

\subsection{Detection Models}
We use RVT-T to process the four event representations in chronological order~\cite{rvt}.
The recurrent vision transformer extracts spatial features from each representation and passes an internal memory state to the next step, combining information across the 200\,ms history.
We initialize this state at the start of each sample and use the final step's detections to localize objects at the annotation time.

YOLOv8n provides a compact frame-based baseline~\cite{yolov8}, and RT-DETR-R18 provides a transformer-based frame comparison~\cite{rtdetr}.
Each receives one image in the corresponding observation condition and predicts boxes for the same five categories.
The comparison therefore evaluates complete observation-and-detector systems: temporally binned event histories and individual original, brightness-scaled, or exposure-integrated frames.
At 200\,ms exposure, frame integration spans the same interval duration as the RVT-T history.

% Reserve the next page top for the main results table.
\begin{table*}[!t]
\centering\small
\caption{Person detection results on held-out source videos, in percent. Best values are bold within each condition and metric.
The 100 ms RVT-T row reuses the original-condition event model and predictions; only the frame inputs change with exposure.
Recall uses confidence 0.25 and matching IoU 0.5; AP$_S$ denotes person AP for small objects.}
\label{tab:main}
\begin{tabular}{llrrrr}
\toprule
Condition & Detector & AP$_{50}$ & AP$_{50:95}$ & Recall & AP$_S$ \\
\midrule
Original & RVT-T &\textbf{37.42}&15.63&37.25&\textbf{4.59}\\
Original & YOLOv8n &32.42&11.73&32.34&3.05\\
Original & RT-DETR-R18 &35.36&\textbf{15.64}&\textbf{37.96}&3.53\\
\midrule
Brightness $\times0.25$ & RVT-T &28.91&13.65&30.18&1.43\\
Brightness $\times0.25$ & YOLOv8n &27.36&11.08&27.31&1.31\\
Brightness $\times0.25$ & RT-DETR-R18 &\textbf{35.81}&\textbf{15.29}&\textbf{36.41}&\textbf{3.92}\\
\midrule
100ms exposure & RVT-T &\textbf{37.42}&\textbf{15.63}&\textbf{37.25}&\textbf{4.59}\\
100ms exposure & YOLOv8n &21.42&6.89&22.16&0.67\\
100ms exposure & RT-DETR-R18 &28.64&9.44&30.42&2.36\\
\bottomrule
\end{tabular}
\end{table*}

\subsection{Training and Evaluation Protocol}
To measure adaptation to each observation condition, we train each frame detector on the image condition used for its evaluation.
All three model families train from scratch on the same recording-level split, with batch size 16, a single training seed, and image augmentation disabled.
Training uses AdamW~\cite{adamw} with cosine learning-rate schedules and runs for approximately 30--90 epochs, with fixed budgets for RVT-T and RT-DETR-R18 and early stopping for YOLOv8n.
Checkpoints are selected using validation performance: five-class AP$_{50:95}$ for RVT-T and RT-DETR-R18, and YOLOv8n's combined AP$_{50}$ and AP$_{50:95}$ fitness.
YOLOv8n uses its built-in training validator for checkpoint selection; the final comparison evaluates every selected model on the common $640\times640$ inputs with the same COCO scoring procedure.
RVT-T and YOLOv8n apply non-maximum suppression, while RT-DETR-R18 uses its native postprocessor.
The main detection experiments use FP32 inference.

We evaluate person detection for the rescue task, using person AP$_{50:95}$ as the primary measure of localization and detection quality.
We also report person AP$_{50}$, precision, recall, and AP$_S$.
All predictions are transformed back to native image coordinates and clipped before evaluation with the same COCO scoring procedure.
Person AP$_S$ is AP$_{50:95}$ for small objects under COCO's native-coordinate area convention; its $32\times32$-pixel area threshold corresponds to roughly $11\times11$ pixels in square-equivalent area after input scaling.

Precision and recall use confidence 0.25 and matching IoU 0.5 throughout.
These rates describe each detector at the specified numerical threshold, with its own confidence-score distribution.
We further evaluate the trained models on an Orin Nano using the deployment and measurement protocol in Section~\ref{sec:deployment}.

% !TeX root = ../main.tex
\section{Results}
We first examine person detection across observation conditions, then study exposure duration and the distribution of detections and false positives across the test scenes.
Table~\ref{tab:main} reports results for the original, reduced-brightness, and 100\,ms exposure conditions.
% Main table is declared near the end of Methodology for placement above these results.

% Declare result figures early enough for the following page.
\begin{figure}[t]
\centering
\includegraphics[width=\linewidth]{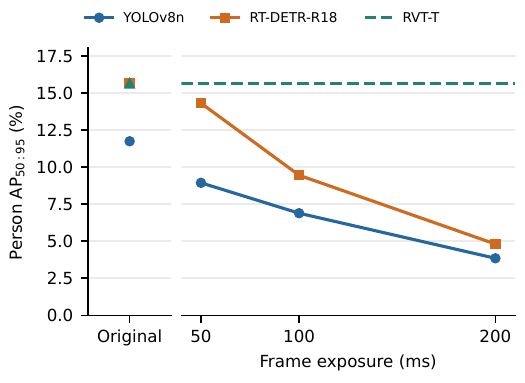}
\caption{Test person AP$_{50:95}$ (\%) as exposure is prolonged from the original video.
Solid curves show exposure-matched frame models; the dashed line shows the same RVT-T predictions reused across exposure conditions.}
\label{fig:exposure}
\end{figure}

\subsection{Person Detection across Observation Conditions}
On the original observations, RVT-T and RT-DETR-R18 achieve nearly identical person AP$_{50:95}$, at 15.63\% and 15.64\%, respectively.

The comparison changes under 100\,ms exposure: RVT-T reaches 15.63\%, while the exposure-matched YOLOv8n and RT-DETR-R18 models reach 6.89\% and 9.44\%.
The event system therefore leads RT-DETR-R18 by 6.19 percentage points in this condition.
Person AP$_{50}$ is 37.42\% for RVT-T, 21.42\% for YOLOv8n, and 28.64\% for RT-DETR-R18, retaining the same ordering.
RVT-T's person AP$_S$ is also higher, at 4.59\% compared with 0.67\% for YOLOv8n and 2.36\% for RT-DETR-R18.
These small-target results are especially relevant to aerial rescue imagery, where a person occupies only a small part of the search scene.

Under reduced brightness, RT-DETR-R18 leads in person AP$_{50:95}$, at 15.29\% compared with 13.65\% for RVT-T and 11.08\% for YOLOv8n.
The clearest event-based benefit in this evaluation is thus person detection in exposure-blurred observations.

\subsection{Exposure Duration}
\begin{figure*}[!t]
\centering
\includegraphics[width=\textwidth]{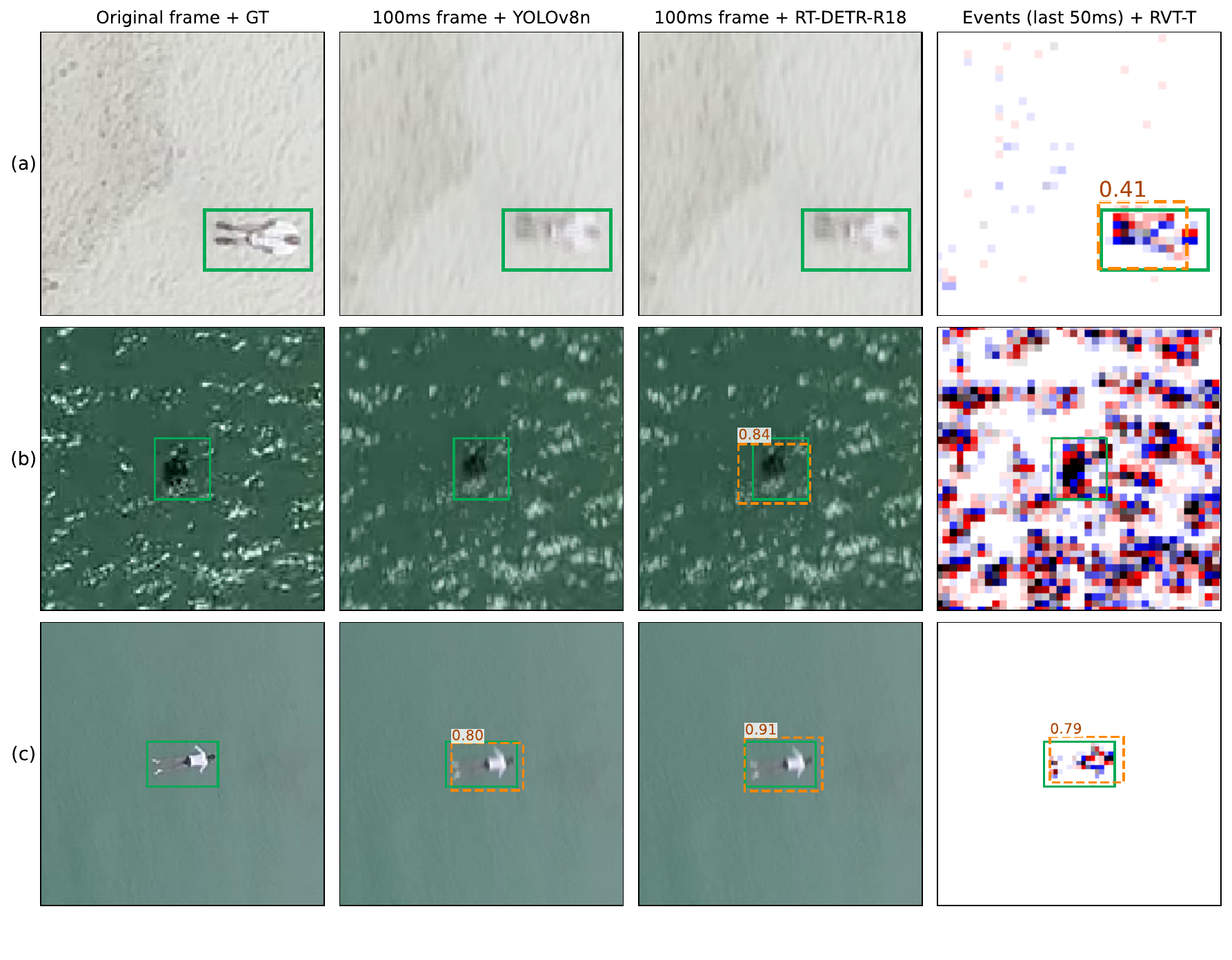}
\caption{Paired person-detection examples from three test recordings, with the same crop across each row.
(a) RVT-T detects the person missed by both frame models.
(b) RVT-T misses the person detected by RT-DETR-R18 amid strong water-background events.
(c) All three models detect the person.
Green boxes show ground truth; dashed orange boxes and scores show predictions at confidence $\geq0.25$.
Frame predictions use 100\,ms exposure images.
Event views show the final 50\,ms, with red/blue for increasing/decreasing brightness, dark for overlapping polarities, and white for no events; RVT-T uses the full 200\,ms history.}
\label{fig:cases}
\end{figure*}

Longer exposure progressively reduces person AP$_{50:95}$ for both frame detectors (Fig.~\ref{fig:exposure}).
Across 50, 100, and 200\,ms, RT-DETR-R18 achieves 14.30\%, 9.44\%, and 4.81\%, while YOLOv8n achieves 8.93\%, 6.89\%, and 3.85\%.
Each frame model is trained for its corresponding exposure condition.
Against the shared source-cadence RVT-T reference of 15.63\%, the gap to RT-DETR-R18 grows from 1.33 to 10.82 percentage points.
As integration duration increases, the frame combines a longer history of target appearances while the annotation identifies the person's position at the end of that interval.

% Declare the two following tables before their discussion to avoid end-of-paper floats.
\begin{table}[t]
\centering\small
\caption{Person precision and recall (\%) for the 100\,ms comparison.}
\label{tab:operating}
\begin{tabular}{lrr}
\toprule
Detector & Precision & Recall\\
\midrule
RVT-T &48.67&37.25\\
YOLOv8n &51.10&22.16\\
RT-DETR-R18 &31.47&30.42\\
\bottomrule
\end{tabular}
\end{table}

% Previous layout retained in place, 2026-10-12.
% \begin{table*}[!t]
% \centering\small
% \caption{Trained Orin Nano deployment with PyTorch CUDA FP16, batch one.
% Person AP$_{50:95}$ (\%) uses all 843 test samples.
% Timing, module-input power and energy report the minimum--maximum across three repetitions.
% The pipeline consumes RAM-resident inputs and includes postprocessing; acquisition and representation construction are excluded.}
% \label{tab:device}
% \begin{tabular}{lrrrrrr}
% \toprule
% Detector & P AP FP32 & P AP FP16 & Pipeline p50 (ms) & Pipeline p95 (ms) & Power (W) & J/sample\\
% \midrule
% RVT-T &15.63&15.63&135.25--136.76&143.23--144.25&14.73--14.77&2.016--2.024\\
% YOLOv8n &6.89&6.75&20.84--21.46&21.43--21.74&10.84--10.89&0.225--0.230\\
% RT-DETR-R18 &9.44&9.50&58.60--59.95&59.02--60.29&12.82--12.95&0.762--0.772\\
% \bottomrule
% \end{tabular}
% \end{table*}

\begin{table*}[!b]
\centering\small
\caption{Detection accuracy and FP16 processing cost on Jetson Orin Nano.}
\label{tab:device}
\setlength{\tabcolsep}{7pt}
\begin{tabular}{lrrrrrr}
\toprule
& \multicolumn{2}{c}{Person AP$_{50:95}$ (\%) $\uparrow$} & \multicolumn{4}{c}{FP16 processing cost}\\
\cmidrule(lr){2-3}\cmidrule(lr){4-7}
Detector & FP32 & FP16 & p50 (ms) $\downarrow$ & p95 (ms) $\downarrow$ & Power (W) & Energy (J/sample) $\downarrow$\\
\midrule
RVT-T &15.63&15.63&135.3--136.8&143.2--144.3&14.73--14.77&2.016--2.024\\
YOLOv8n &6.89&6.75&20.8--21.5&21.4--21.7&10.84--10.89&0.225--0.230\\
RT-DETR-R18 &9.44&9.50&58.6--59.9&59.0--60.3&12.82--12.95&0.762--0.772\\
\bottomrule
\end{tabular}
\end{table*}

\subsection{Scene Variation and False Positives}
% Source table and its original discussion retained in place for restoration, 2026-10-12.
% \input{tables/person-sources}
% The 100\,ms comparison varies substantially across the five test recordings (Table~\ref{tab:sources}).
% All three systems have near-zero person AP on source A, while sources D and E account for much of the pooled performance.
% RVT leads both frame models on the other four sources.
% Recomputing pooled AP after omitting each source retains the RVT lead in all five cases.
% The low scores on several recordings identify scene-dependent detection as an important issue for further development.
The 100\,ms comparison varies substantially across the five test recordings, with all three systems obtaining near-zero person AP$_{50:95}$ on one recording.
RVT-T leads both frame models on the other four recordings, and its pooled person AP$_{50:95}$ remains higher after omitting any one recording.

% Operating-point table is declared immediately before this subsection.
At confidence 0.25 and matching IoU 0.5, RVT-T has person recall of 37.25\%, compared with 22.16\% for YOLOv8n and 30.42\% for RT-DETR-R18 (Table~\ref{tab:operating}).
Relative to YOLOv8n, this higher recall accompanies slightly lower precision, 48.67\% versus 51.10\%.
RT-DETR-R18 has precision of 31.47\% at this operating point.
Higher recall means that more annotated people are detected, while lower precision means that a larger fraction of candidate detections are false positives. For rescue teams, the gain in person detection therefore comes with additional false detections to review.

% Qualitative figure declared earlier for placement beside the results.

Figure~\ref{fig:cases} illustrates the visual conditions behind the comparison.
% In the first row, the upper person's appearance is spread by exposure integration, while the event representation contains a compact response and RVT detects the target.
In the first row, the person's appearance is spread by exposure integration, while the event representation contains a compact response and RVT-T detects the target.
In the second row, water activity produces a strong event background and RVT-T misses a person detected by RT-DETR-R18.
All three models detect the person in the third row.
Together, these examples show the value of retaining target detail during motion and the continuing importance of separating a small person from the surrounding water.

% !TeX root = ../main.tex
\section{Embedded Deployment}
\label{sec:deployment}
The detection benefit under exposure blur motivates examining the computing resources needed to use the event model onboard a search UAV.
We deploy source-cadence RVT-T and the two 100\,ms-trained frame models on a Jetson Orin Nano and compare their accuracy, latency, and energy per sample.
% Device table is declared in the preceding section to appear beside this discussion.

\subsection{Deployment and Measurement}
All models use batch-one PyTorch CUDA inference on the same device in MAXN\_SUPER mode with locked CPU, GPU, and memory clocks.
FP32 and FP16 accuracy are evaluated on all 843 test samples.
Latency and energy measurements use a shared set of 16 test samples held in RAM and sampled across the test set.
After warmup, each measurement runs for at least 1,000 sample evaluations and 60 seconds.
We repeat the measurements three times and rotate model order between repetitions.

The measured pipeline begins with prepared event histories or resized images in memory and ends with detections in native image coordinates, including CPU-to-GPU transfer, dtype conversion, model inference, and postprocessing.
Each RVT-T sample transfers its full 200\,ms history, resets the recurrent state, and processes all four 50\,ms windows.
Input preparation, acquisition, and disk access lie outside this timing interval.
We integrate module-input power over each run and divide by the number of processed samples to obtain energy per sample, including idle consumption.
Table~\ref{tab:device} reports the range across the three FP16 repetitions for median (p50) and 95th-percentile (p95) latency, mean module-input power, and energy.

\subsection{Accuracy and Processing Cost}
FP16 preserves person AP$_{50:95}$ to within 0.14 percentage points of FP32 for all three models.
The deployed RVT-T therefore retains its detection advantage in the 100\,ms comparison, with person AP$_{50:95}$ of 15.63\% versus 6.75\% for YOLOv8n and 9.50\% for RT-DETR-R18.
The higher person detection performance comes with a higher processing cost: RVT-T requires approximately 136\,ms and 2.02\,J per sample, compared with 21\,ms and 0.23\,J for YOLOv8n and 59\,ms and 0.77\,J for RT-DETR-R18.

The current RVT-T implementation processes event-count representations with a transformer on the GPU.
Spiking neural networks (SNNs) paired with neuromorphic processors~\cite{neuromorphicdrone} offer a further direction for reducing energy through sparse, event-driven computation.

% !TeX root = ../main.tex
\section{Discussion and Conclusion}
This study examines event-based person detection for maritime UAV search using simulated events from annotated MOBDrone footage.
The comparison with frame-based detectors shows a person-detection advantage under synthetic exposure blur, making event sensing a promising option for observing small targets during camera motion.
Video-to-event conversion allows this application to be explored using existing rescue recordings and annotations.

Separating a person from activity on the water remains central to the monitoring task.
The event representation can retain a compact target response in blurred scenes, while water motion also produces events that compete with the person.
For a search operator, useful detections should draw attention to a possible person without repeatedly flagging the surrounding water.
Following candidate detections over time and combining event observations with conventional imagery are practical directions for improving this distinction.

A next step is to evaluate synchronized frame and event recordings of people in the water under different sea states, illumination levels, and camera motions.
This would allow the detection performance and false positives observed here to be examined during continuous search sequences, together with the processing demands on the UAV.

\section*{Acknowledgement}
This research was funded by the Flemish Government (AI Research Program).
We acknowledge EuroHPC Joint Undertaking for awarding us access to LEONARDO, hosted by CINECA, Italy.

{
    \small
    \bibliographystyle{ieeenat_fullname}
    \bibliography{references}
}
\end{document}